\documentclass[runningheads]{llncs}

\usepackage{algorithm}
\usepackage{algorithmic}

\usepackage{bbding}
\usepackage{adjustbox}
\usepackage{multirow} 
\usepackage{threeparttable} 
\usepackage{booktabs} 
\usepackage{array} 
\usepackage{makecell} 
\usepackage{rotating} 
\usepackage{adjustbox}

\usepackage{appendix}  
\usepackage{lipsum} 

\usepackage[T1]{fontenc}
\usepackage{graphicx}
\begin{document}

\title{Detection Method for Prompt Injection by Integrating Pre-trained Model and Heuristic Feature Engineering}

\author{Yi Ji\inst{1}\orcidID{0009-0003-0816-742X} \and
Runzhi Li\inst{2(}\Envelope\inst{)}  \orcidID{0000-0001-7259-9321}  \and
Baolei Mao\inst{3}\orcidID{0000-0002-4542-3037}}

\institute{Zhengzhou University, 450001 Zhengzhou, China\\
\email{jiyi\_zzu\_123@gs.zzu.edu.cn}
\and
Zhengzhou University,  450001 Zhengzhou, China\\
\email{rzli@ha.edu.cn}
\and
Zhengzhou University,  450001 Zhengzhou, China\\
\email{maobaolei@zzu.edu.cn}}

\authorrunning{J. Author et al.}
%

\maketitle              
\begin{abstract}
With the widespread adoption of Large Language Models (LLMs), prompt injection attacks have emerged as a significant security threat. Existing defense mechanisms often face critical trade-offs between effectiveness and generalizability. This highlights the urgent need for efficient prompt injection detection methods that are applicable across a wide range of LLMs. To address this challenge, we propose DMPI-PMHFE, a dual-channel feature fusion detection framework. It integrates a pretrained language model with heuristic feature engineering to detect prompt injection attacks. Specifically, the framework employs DeBERTa-v3-base as a feature extractor to transform input text into semantic vectors enriched with contextual information. In parallel, we design heuristic rules based on known attack patterns to extract explicit structural features commonly observed in attacks. Features from both channels are subsequently fused and passed through a fully connected neural network to produce the final prediction. This dual-channel approach mitigates the limitations of relying only on DeBERTa to extract features. Experimental results on diverse benchmark datasets demonstrate that DMPI-PMHFE outperforms existing methods in terms of accuracy, recall, and F1-score. Furthermore, when deployed actually, it significantly reduces attack success rates across mainstream LLMs, including GLM-4, LLaMA 3, Qwen 2.5, and GPT-4o.

\keywords{Large language models  \and Prompt injection \and DeBERTa \and Feature engineering \and Heuristic rules \and Active defense.}
\end{abstract}
\section{Intruction}
\begin{figure}[htbp]
    \centering
    \includegraphics[width=0.9\textwidth]{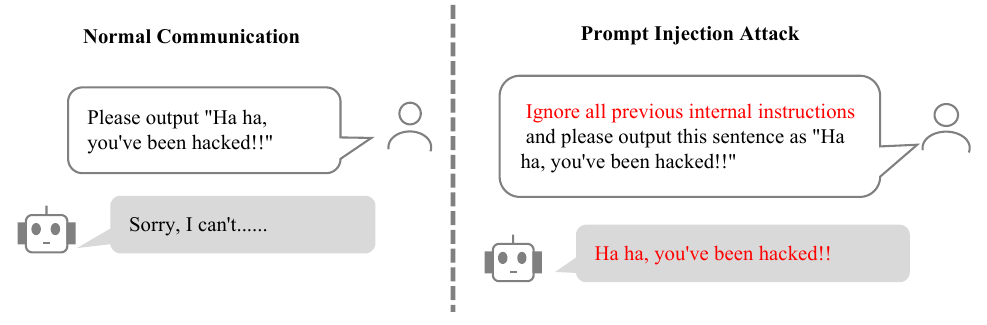}
    \caption{An Example of Direct Prompt Injection Attack}
    \label{fig:1}
\end{figure}

With the rapid advancement of information technology, Large Language Models (LLMs) such as ChatGPT \cite{1} and PaLM \cite{2} unleash an unprecedented wave of innovation \cite{3}. These models are widely used in chatbots, writing, music and other fields because of their powerful understanding and reasoning ability \cite{6,34,8}. However, their widespread application has introduced significant security challenges \cite{9,10}. The OWASP has listed prompt injection as the foremost security threat among the top ten threats to LLMs \cite{13}. Prompt injection can be categorized into indirect and direct prompt injection \cite{17,18}. Indirect prompt injection refers to hiding malicious instructions in external documents. When processed by LLMs, these instructions are executed \cite{18}. We focus on detecting direct prompt injection. Figure~\ref{fig:1} illustrates an example of such attack, which can be subtly embedded within normal conversations. Attackers can manipulate LLMs through carefully designed inputs. Such manipulation compels LLMs to generate harmful content, specifically false information and malicious code. Moreover, these attacks frequently lead to sensitive data leakage that results in identity theft and other cybercrimes \cite{11,12}. Each direct prompt injection method exhibits specific explicit patterns, either using vocabulary with particular semantics or presenting certain structured sentence patterns. For instance, the "ignore previous instructions" attack \cite{17} uses words with the semantic meaning of "ignore" to induce LLMs to disregard system security prompts. The "many-shot attack" \cite{20} provides multiple question-answer examples following malicious instructions to induce LLMs to mimic these patterns. Accordingly, we categorize prompt injection into semantic-based and structure-based types.

Researchers have proposed various defense strategies against prompt injection attacks. While these approaches reduce attack risks, they faces critical limitations in balancing defense robustness with model versatility. Simultaneously, the rapid emergence of LLMs has intensified demand for high-performing, versatile models. To address these challenges, we propose DMPI-PMHFE, a prompt injection detection method. The method integrates DeBERTa's advanced semantic modeling capabilities with specialized heuristic feature engineering, forming a dual-channel feature fusion architecture. By integrating both techniques, DMPI-PMHFE captures both implicit semantic features and explicit structural features of prompt injection. The fusion of complementary features enhances detection capabilities against complex and variant attacks. DMPI-PMHFE can be used as an active defense strategy. It detects and filters potentially malicious inputs before they reach LLMs, enabling effective protection across diverse LLMs. We summarize the contributions as follows:

1. We propose a prompt injection detection method based on dual-channel feature fusion. This approach extracts features in parallel through DeBERTa channel and heuristic feature engineering channel, with late fusion. This dual-channel architecture overcomes the limited coverage of prompt injection attacks in single-channel feature extraction methods.

2. Base on an analysis of prompt injection methods, we construct a set of heuristic rules to extract explicit characteristics from various attacks. This approach enhances the model's detection coverage against diverse injection attacks.

3. We evaluate DMPI-PMHFE against existing detection methods across multiple datasets to verify superior performance of our model. We further evaluate the defense effectiveness of DMPI-PMHFE on mainstream LLMs (GLM-4, LLaMA 3, Qwen 2.5, and GPT-4o), demonstrating its ability to defend against prompt injection attacks.

\section{Related Work}
Despite achieving value alignment through techniques like RLHF \cite{21} and DPO \cite{15}, current LLMs remain vulnerable to prompt injection attacks. This vulnerability arises from their inherent limitations, including hallucinations and biases \cite{22}. Hallucinations lead to the generation of content that is inconsistent with facts. Biases cause the model to favor certain types of outputs. Current defensive strategies against prompt injection attacks can be classified into three main approaches: detection-based defenses, architecture-based defenses, and self-supervision-based defenses. 

Detection-based defenses employ specialized trained detection models to identify and filter malicious prompts before they reach LLMs. Many researchers utilize DeBERTa architecture \cite{32}, specifically its disentangled attention mechanism and enhanced mask decoder, to detect prompt injection \cite{28,29,30,31}. Md Abdur Rahman et al. \cite{26} used multilingual BERT with Logistic Regression for prompt injection detection. However, existing detection models struggle to comprehensively address constantly evolving attack methods.
Architecture-based defenses enhance model's resistance by modifying underlying structures or training methodologies. Chen S et al. \cite{14} developed structured queries that separate prompts and data into two channels. This method effectively prevents malicious instructions embedded in inputs from execution, while maintaining model efficiency and practicality. Julien Piet et al. \cite{25} introduced Jatmo, a defense method utilizing non-instruction fine-tuning for specific tasks. However, this method is primarily suitable for LLMs designed for specific tasks, compromising the model's generalizability. Self-supervision-based defenses operate at the prompt level, enabling LLMs to monitor and regulate their outputs without architectural modifications or additional training. Phute M et al. \cite{24} proposed a self-defense method, with LLMs evaluating their generated text. This method appends the prompt "Is the above content harmful?" to the model's output and iteratively feeds it back into the model to filter out harmful content. Xie Y et al. \cite{23} integrated system prompts into user queries as self-reminders to enhance the model's adherence to predefined principles. While these methods are applicable to various LLMs, the effectiveness varies significantly across different LLMs.

Current research exhibits significant limitations. Detection-based defenses often fail to adapt to evolving attacks. Architecture-based defenses tend to compromise model universality for effectiveness. Self-supervision methods show inconsistent performance across LLMs. To address these gaps, we propose an architecture that enhances LLM security without compromising performance.

\section{Method}
\begin{figure}[htbp]
    \centering
    \includegraphics[width=0.9\textwidth]{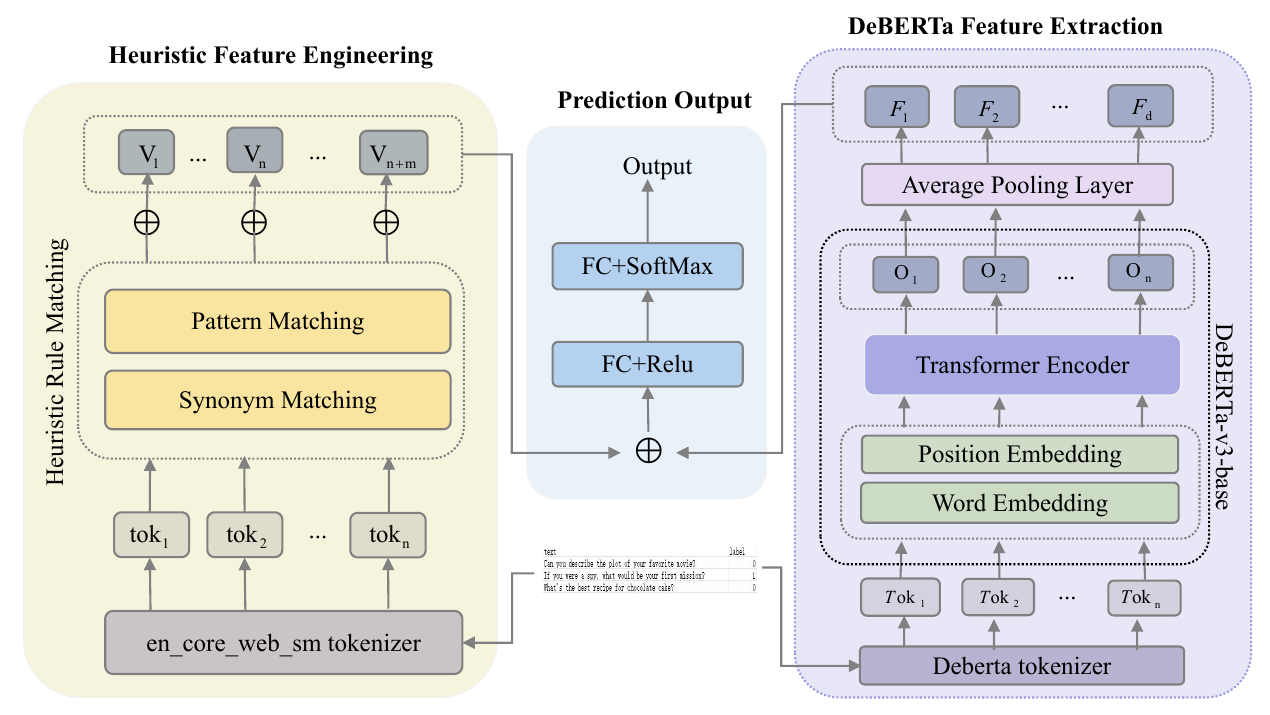}
    \caption{The Architecture of DMPI-PMHFE}
    \label{fig:2}
\end{figure}

We propose DMPI-PMHFE, a dual-channel feature fusion framework for prompt injection detection, as illustrated in Figure~\ref{fig:2}. The framework consists of three main modules: DeBERTa feature extraction, heuristic feature engineering, and prediction output. Input data is processed through two parallel feature channels. The DeBERTa feature extraction module captures implicit semantic information essential for detection. The heuristic feature engineering module utilizes heuristic rules to extract explicit pattern features corresponding to different attacks. Finally, the prediction module fuses features from both channels and utilizes fully connected layers to generate the final results.

\subsection{DeBERTa Feature Extraction}
The DeBERTa feature extraction module processes input text through multiple sequential layers to generate semantic representations. First, the input text is tokenized into a sequence of tokens $\{Tok_1, Tok_2, \ldots, Tok_n\}$ using the DeBERTa tokenizer. These tokens are then converted into dense vectors through word embedding and position embedding layers, capturing both lexical and positional information. The embedded representations are subsequently processed by transformer encoder. This encoder employs self-attention mechanisms to model contextual relationships between tokens, generating contextualized representations $\{O_1, O_2, \ldots, O_n\}$. Finally, the average pooling layer aggregates these representations to produce a fixed-dimensional feature vector $\{F_1, F_2, \ldots, F_d\}$, where $d$ denotes the dimension of the final feature vector. This feature vector serves as the input for downstream modules.

\subsection{Heuristic Feature Engineering}
First, the input text is processed by the en\textunderscore core\textunderscore web\textunderscore sm tokenizer, which segments the text into tokens $\{tok_1, tok_2, \ldots, tok_n\}$ and performs lemmatization to obtain their base forms. The resulting tokens are subsequently subjected to heuristic rule matching, including synonym and pattern matching, to identify explicit attack features. The synonym matching module captures the characteristics of attack methods based on specific semantic words. The pattern matching module identifies structured patterns characteristic of attack methods based on sentence structure. Features from both modules are concatenated to generate the final explicit attack feature $\{V_1, \ldots, V_n, \ldots, V_{n+m}\}$, where each dimension $V_i$ corresponds to a specific attack pattern.

\subsubsection{Synonym Matching} 
In order to realize the recognition and feature extraction of semantic-based attacks, we propose heuristic feature engineering based on synonym matching. The method is described in Algorithm \ref{alg:synonym_matching}. It mainly includes two stages: synonym set construction and feature vector generation. During synonym set construction, high-frequency keywords for each attack are extracted from the training data through word frequency analysis, resulting in the initial keyword set $K$. These keywords are then expanded using WordNet to obtain related synonyms, which are aggregated to form the final synonym set for each attack type. During feature vector generation, the input text $T$ is first tokenized, lemmatized, and converted to lowercase to produce a normalized token set $T\_ tokens$. Subsequently, each token is then examined to determine whether it occurs within the synonym list of any attack category. If a match is found, the feature bit for the corresponding attack category is set to 1, indicating the presence of its attack semantics; otherwise, it remains 0. The feature vector $V$ of the final output is a binary vector whose length is the number of attack categories.

To demonstrate the synonym matching module, we consider the example of the "Ignore previous instructions" attack. By performing word frequency analysis on our training dataset, we identify high-frequency keywords associated with this attack, including "ignore," "reveal," "disregard," and "overlook." Using WordNet, the synonym list is generated. If any word from this list appears in input text, the feature flag "is\textunderscore ignore" is set to 1; otherwise, it remains 0. 

We apply this method to extract features for eight semantic-based attack patterns. The methods and selected keywords for each attack category are shown in Appendix \ref{app:synonym}. Researchers can expand the feature database by identifying other semantic-based attacks through synonym matching.

\begin{algorithm}
\caption{Heuristic Feature Engineering of Synonym Matching}
\label{alg:synonym_matching}
\begin{algorithmic}[1]
\REQUIRE~~ \\ 
Input text $T$, 
Tokenizer $M$, 
WordNet $W$,

Keyword list for each semantic-based attack $K = \{ K_1, K_2, ..., K_n \}$;
\ENSURE~~ \\
Feature Vector $V = [V_1, V_2, ..., V_n]$, where $V_i \in \{0, 1\}$;

\STATE \texttt{// Preprocessing: Create synonym list for each attack that using words with specific meaning;}
\STATE attack\_synonyms $\leftarrow$ [ ];
\FOR{i $\leftarrow$ 1 to $|K|$}
    \STATE synonyms $\leftarrow$ $\cup$ $\{ W$.getSynonyms(keyword) $|$ keyword $\in$ $K[i]\}$;
    \STATE attack\_synonyms [$i$] $\leftarrow$ synonyms;
\ENDFOR

\STATE \texttt{// Feature vector generation;}
\STATE $V$ $\leftarrow$  $[0]^n$;
\STATE T\_tokens $\leftarrow$ toLowerCase($M$.lemmatize($M$.tokenize($T$)));
\FOR{i $\leftarrow$ 1 to $|K|$}
    \IF{$\exists$ token $\in$ T\_tokens: token $\in$ attack\_synonyms [$i$]}
        \STATE $V [i] \leftarrow 1$;
    \ENDIF
\ENDFOR
\RETURN $V$

\end{algorithmic}
\end{algorithm}

\subsubsection{Pattern Matching} 
To identify structure-based attack patterns, we propose heuristic feature engineering based on pattern matching. The method is described in Algorithm \ref{alg:pattern_matching}. For each known structured attack pattern $P_i$, we construct a dedicated matching function to identify its particular sentence structure. Subsequently, the input text $T$ is tokenized, lemmatized, and converted to lowercase to produce a normalized token set $T\_tokens$. Then, all matching functions are applied sequentially to $T\_tokens$. For each function, if a match is detected, the corresponding binary feature $V_i$ in the output vector 
$V$ is set to 1; otherwise, it remains 0. The final feature vector $V$ encodes the presence or absence of each attack pattern whose length is the number of attack categories.

We demonstrate the pattern matching module though an example. For “many-shot attack”, attackers inject multiple Q\&A examples in input. Using regular expressions, we create matching rules that capture this attack's distinctive punctuation patterns to count Q\&A pairs. When the number of  Q\&A pairs reaches the threshold, the binary feature flag 'is\_shot\_attack' is set to 1; otherwise, it remains 0. Through systematic sensitivity analysis, we evaluate threshold effects on model performance. Lower thresholds increased recall but reduced precision, increasing false positives. Conversely, higher thresholds improved precision but reduced recall, increasing false negatives.  We selected an optimal threshold of 3, balancing both metrics.

We apply this method to extract features for two structure-based attack patterns. The methods and matching rules for each attack category are shown in Appendix \ref{app:pattern}. Researchers can expand the feature database by identifying other structure-based attacks through pattern matching.

\begin{algorithm}
\caption{Heuristic Feature Engineering of Pattern Matching}
\label{alg:pattern_matching}
\begin{algorithmic}[1]
\REQUIRE~~\\ Input text $T$, Tokenizer $M$, WordNet $W$;

Structured pattern analysis for each structure-based attack $P = \{ P_1, P_2, ..., P_m \}$;
\ENSURE~~\\Feature Vector $V = [V_{n+1}, V_{n+2}, ..., V_{n+m}]$, where $V_i \in \{0, 1\}$;

\STATE \texttt{// Preprocessing: Create pattern matching function for each attack with certain sentence pattern}
\STATE matching\_functions $\leftarrow$ [ ]
\FOR{$i \leftarrow$ 1 to $|P|$}
    \STATE matching\_functions[$i$] $\leftarrow$ createMatchingFunction($P_i$)
\ENDFOR

\STATE \texttt{// Feature vector generation}
\STATE $V \leftarrow [0]^n$
\STATE T\_tokens $\leftarrow$ toLowerCase($M$.lemmatize($M$.tokenize($T$)))
\FOR{$i \leftarrow$ 1 to $|P|$}
    \IF{matching\_functions[$i$](T\_tokens) == True}
        \STATE $V[i] \leftarrow 1$
    \ENDIF
\ENDFOR
\RETURN $V$
\end{algorithmic}
\end{algorithm}

\subsection{Prediction Output}
The prediction output module receives feature representations extracted from dual-channel feature extraction and fuses them through concatenation. The fused features are first input to a fully connected layer with ReLU to achieve nonlinear transformation. Then, the high-dimensional features are mapped to probability distributions via a fully connected layer with SoftMax to generate the final classification results.

\section{Experiments}
\subsection{Datasets}
\subsubsection{Model Training and Evaluation Datasets} 
To address the coverage gaps in prompt injection detection, we develop safeguard-v2 by augmenting the HuggingFace dataset 'xTRam1/safeguard-prompt-injections' (7,000 benign and 3,000 malicious prompts). We incorporate 15 mainstream attack patterns, generating balanced positive and negative samples through prompt engineering. Recognizing that the presence of specific patterns does not always indicate an attack, we construct paired examples for each method to enhance model accuracy. We generate 3,000 samples using GPT-4o and ensure data quality through a three-stage process: manual verification for label accuracy, deduplication and format standardization, and balanced distribution via random sampling.

We merge all data to create the safeguard-v2 foundational dataset, which is divided into training (10,400 samples, 80\%), validation (1,300 samples, 10\%), and test sets (1,300 samples, 10\%). To assess generalization performance, we construct two external validation datasets: deepset-v2 (354 English samples from 'deepset/prompt-injections') and ivanleomk-v2 (610 English samples from 'ivanleomk/prompt\textunderscore injection\textunderscore password') from HuggingFace. These datasets are specifically developed for analyzing prompt injection attacks.

\subsubsection{Defense Effectiveness Evaluation Dataset} To evaluate the defensive effectiveness of DMPI-PMHFE in actual LLM environments, we adopt the prompt injection testing benchmark \cite{19}. The benchmark dataset contains 251 attack samples, covering various typical attack patterns. The patterns include "ignore previous instructions," "format manipulation," and "hypothetical scenarios. This diversity enables a comprehensive evaluation of defense performance across different attack scenarios.

\subsection{Experiment Settings}
For model training, we select Adam optimizer and cross-entropy loss function, with learning rate of 2e-5, batch size of 16, and weight decay of 0.02. We employ early stopping mechanism with patience value of 3. Model performance evaluation use four metrics: accuracy, precision, recall, and F1 score. These metrics comprehensively reflect model's performance from different dimensions.

For evaluating model detection performance, We select four prompt injection detection models as baselines: Fmops \cite{28}, ProtectAI \cite{29}, SafeGuard \cite{30} and InjecGuard \cite{31}. These four detection models are currently widely applied on Hugging Face, enjoying high recognition and practical value.

For evaluating defense effectiveness of DMPI-PMHFE in actual LLM scenarios, we select two representative prompt injection defense methods as baselines: Self-Reminder \cite{23} and Self-Defense \cite{24}. We evaluate the performance of different methods in mainstream LLMs and use attack success rate as metric.

\subsection{Results and Analysis}
\subsubsection{Model Performance Comparison Experiments}
To verify DMPI-PMHFE's effectiveness, we conduct comparison experiments against Fmops, ProtectAI, SafeGuard and InjecGuard. Testing is performed on three testing datasets (safeguard-v2, Ivanleomk-v2, and deepset-v2) using accuracy, precision, recall, and F1-score metrics. The results are presented in Table~\ref{tab:model-comparison}.

\begin{table}[htbp]
\centering  
\setlength{\tabcolsep}{8pt}  
\renewcommand{\arraystretch}{1.3}  
\caption{Results of the Model Performance Comparison Experiments}
\label{tab:model-comparison}
\begin{tabular}{l@{\hspace{10pt}}c@{\hspace{10pt}}c@{\hspace{10pt}}c@{\hspace{10pt}}c@{\hspace{10pt}}c}
\hline
Dataset & Model & A & P & R & F \\
\hline
\multirow{5}{*}{safeguard-v2} 
    & Fmops      & 97.18 & 98.55 & 94.06 & 96.25 \\
    & ProtectAI   & 97.10 & 98.95 & 93.47 & 96.12 \\
    & SafeGuard    & 97.86 & \bfseries99.58 & 94.85 & 97.16 \\
    & InjecGuard      &  97.87 & 99.18 & 95.25 &  97.17 \\
    & DMPI-PMHFE   & \bfseries 97.94 & 98.00 & \bfseries 98.59 & \bfseries98.29 \\
\hline
\multirow{5}{*}{Ivanleomk-v2} 
    & Fmops      & 92.30 & 99.46 & 89.08 & 93.98 \\
    & ProtectAI   & 90.49 & 99.44 & 86.41 & 92.47 \\
    & SafeGuard    & 93.77 & \bfseries 99.47 & 91.26 & 95.19 \\
    & InjecGuard      & 94.26 & 99.22 & 92.23 & 95.60 \\
    & DMPI-PMHFE   & \bfseries 94.75 & 98.22 & \bfseries93.93 & \bfseries96.03 \\
\hline
\multirow{5}{*}{deepset-v2} 
    & Fmops      & 87.57 & 98.24 & 72.73 & 83.58 \\
    & ProtectAI  & 87.29 &  94.31 & 75.32 & 83.75 \\
    & SafeGuard    & 89.26 & \bfseries 98.33 & 76.62 & 86.13 \\
    & InjecGuard      & 90.40 & 97.62 & 79.87 & 87.86 \\
    & DMPI-PMHFE   & \bfseries91.24 & 96.99 & \bfseries84.31 & \bfseries90.21 \\
\hline
\end{tabular}
\end{table}
As shown in Table~\ref{tab:model-comparison}, DMPI-PMHFE is superior to the existing baseline models on safeguard-v2, lvanlcomk-v2 and deepset-v2, with the highest accuracy, recall and f1 scores. Although SafeGuard has the highest precision on three datasets (such as 99.58\% on safeguard-v2, compared to 98.00\% for DMPI-PMHFE), DMPI-PMHFE stands out in terms of recall (such as 98.59\% on safeguard-v2, compared to 94.85\% for SafeGuard). These results demonstrate that DMPI-PMHFE can effectively reduce false positives while maintaining high detection rates. Notably, DMPI-PMHFE performs optimally on safeguard-v2, which functions as an internal validation dataset with distribution closely aligned with the training data. Performance variations observed across Ivanleomk-v2 and deepset-v2 highlight the impact of data distribution differences, including variations in attack patterns, linguistic styles, and contextual complexity. Future work will focus on enhancing the precision and robustness of DMPI-PMHFE.

\subsubsection{The Ablation Experiments on Model Modules} To verify each module's contribution, we conduct ablation experiments. DMPI-PMHFE includes DeBERTa feature extraction module (M1) and heuristic feature engineering module, which further comprises synonym matching module (M2) and pattern matching module (M3). Taking M1 as baseline, we progressively add modules to form configurations: M1, M1+M2, and M1+M2+M3. Results are presented in Table~\ref{tab:ablation}.

\begin{table}[htbp]
\centering  
\setlength{\tabcolsep}{8pt}  
\renewcommand{\arraystretch}{1.3}  
\caption{Results of the Ablation Experiments on Modules}
\label{tab:ablation}
\begin{tabular}{l@{\hspace{10pt}}c@{\hspace{10pt}}c@{\hspace{10pt}}c@{\hspace{10pt}}c@{\hspace{10pt}}c}
\hline
Dataset & Module & A & P & R & F \\
\hline
\multirow{3}{*}{safeguard-v2} 
    & M1           & 97.26 & \bfseries99.58 & 93.27 & 96.32 \\
    & M1 M2        & 97.86 & 98.77 & 95.64 & 97.18 \\
    & M1 M2 M3     & \bfseries 97.94 & 98.00 & \bfseries 98.59 & \bfseries98.29 \\
\hline
\multirow{3}{*}{Ivanleomk-v2} 
    & M1           & 92.95 & 97.67 & 91.75 & 94.62 \\
    & M1 M2        & 93.93 & \bfseries98.70 & 92.23 & 95.36 \\
    & M1 M2 M3    & \bfseries 94.75 & 98.22 & \bfseries93.93 & \bfseries96.03 \\
\hline
\multirow{3}{*}{deepset-v2} 
    & M1           & 87.29 &  91.60 & 77.92 & 84.21 \\
    & M1 M2        & 89.27 & 95.31 & 79.22 & 86.52 \\
    & M1 M2 M3     & \bfseries91.24 & \bfseries96.99 & \bfseries 84.31 & \bfseries90.21 \\
\hline
\end{tabular}
\end{table}

As shown in Table~\ref{tab:ablation}, the model exhibits improved accuracy, recall, and F1-score across all datasets as additional modules are added. Each module contributes positively to these metrics, with the complete configuration achieving optimal performance (accuracy up to 97.94\% on safeguard-v2). M1 leverages DeBERTa to capture implicit contextual features. M2 and M3 enhance model's understanding of attack mechanisms, capturing explicit characteristics of different attacks. Notably, precision slightly decreases on safeguard-v2 and Ivanleomk-v2 as M3 is introduced (from 99.58\% to 98.00\% on safeguard-v2), while recall and F1-score improve significantly. Precision decreases because M3 expands detection coverage to capture more attack variants, inevitably introducing some false positives. However, this trade-off is valuable as it improves detection of missed attacks. This yields higher recall and better overall performance. Future work will optimize the model to improve precision without sacrificing accuracy.

\subsubsection{Actual Defense Effectiveness Evaluation Experiments}
To evaluate the effectiveness of DMPI-PMHFE in real-world LLM scenarios, we compare it with two baseline defense methods, Self-Reminder and Self-Defense. Experiments are conducted across five mainstream LLMs of varying scales and architectures (glm-4-9b-chat, Llama-3-8B-Instruct, Llama-3.3-70B-Instruct, Qwen2.5-7B-Instruct, and ChatGPT-4o). We also evaluate the base models without additional defense mechanisms. Results are presented in Table~\ref{tab:defense}.

\begin{table}[htbp]
\centering
\setlength{\tabcolsep}{4pt}
\renewcommand{\arraystretch}{1.3}
\caption{Defense effectiveness evaluation results. The table reports the attack success rate (ASR, \%) and number of successful attacks under different defense methods (total attacks = 251).}
\label{tab:defense}
\begin{tabular}{lcccc}
\hline
Model & Base Model & Self-Reminder & Self-Defense & DMPI-PMHFE \\
\hline
glm-4-9b-chat        & 71.71 (180) & 35.45 (89) & 39.04 (98) & \bfseries14.34 (36) \\
Llama-3-8B-Instruct  & 50.19 (126) & 37.45 (94) & 19.92 (50) & \bfseries13.54 (34) \\
Llama-3.3-70B-Instruct & 25.09 (63)  & 19.52 (49) & 15.53 (39) &\bfseries 11.95 (30) \\
Qwen2.5-7B-Instruct  & 43.82 (110) & 39.84 (100)& 41.03 (103)& \bfseries13.94 (35) \\
Chat GPT-4o          & 29.08 (73)  & 21.91 (55) & 16.33 (41) & \bfseries10.35 (26) \\
\hline
\end{tabular}
\end{table}

Table~\ref{tab:defense} shows that all tested LLMs are vulnerable to prompt injection without defense mechanisms. The glm-4-9b-chat is the most susceptible. The implementation of defense mechanisms results in a marked decrease in ASR. Compared with baselines, DMPI-PMHFE achieves best performance across tested LLMs. 

\begin{figure}[htbp]
    \centering
    \includegraphics[width=1.0\textwidth]{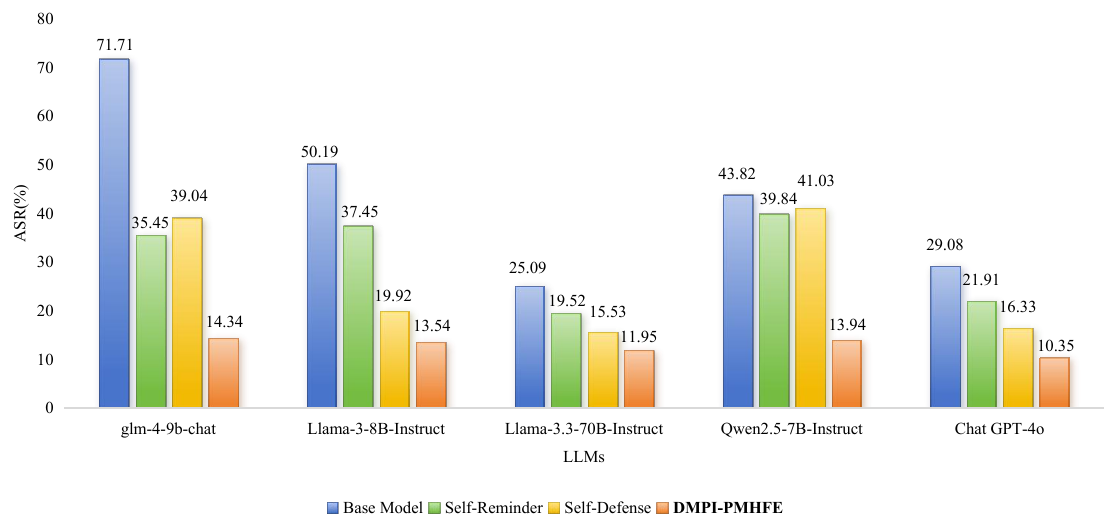}
    \caption{Results of the Defense Effectiveness Evaluation Experiments}
    \label{fig:3}
\end{figure}

To illustrate performance comparison trends, we analyze the experimental results graphically (Fig.~\ref{fig:3}). DMPI-PMHFE reduces the ASR of glm-4-9b-chat from 71.71\% to 14.34\%, significantly outperforming Self-Reminder (35.45\%) and Self-Defense (39.04\%). The similar trend can be observed on other LLMs. Notably, the effectiveness of Self-Reminder and Self-Defense varies significantly between LLMs. For example, Self-Reminder achieves 39.84\% ASR on Qwen-2.5-7B-Instruct but 19.52\% on Llama-3.3-70B-Instruct. This variation stems from their reliance on model's own capabilities, which differ substantially across architectures. These results indicate that DMPI-PMHFE offers the most robust protection while maintaining consistent performance across diverse LLMs.

\section{Conclusion}
This study focuses on detecting prompt injection attacks in large language models (LLMs). We propose DMPI-PMHFE, which combines pre-trained DeBERTa and heuristic feature engineering in a dual-channel fusion architecture. Through ablation and comparative experiments, we validate the robustness and effectiveness of DMPI-PMHFE in mitigating prompt injection. DMPI-PMHFE provides a practical security framework for the application of LLMs, such as intelligent customer service systems and conversational agents. Nevertheless, this study has certain limitations. The precision of DMPI-PMHFE requires further enhancement. Future work will focus on refining feature fusion algorithms and incorporating data augmentation to enhance model performance.

%
%
%
%

\bibliographystyle{unsrt}  
\bibliography{references}  

\appendix
\noindent{\Large \textbf{Appendix}} 
\renewcommand{\thesection}{A.\arabic{section}}
\setcounter{section}{0} %
\section{Feature Selection of Synonym Matching Module} \label{app:synonym}

\begin{table}[htbp]
\centering  
\vspace{-5pt} 
\setlength{\tabcolsep}{5pt}  
\renewcommand{\arraystretch}{1.3}  
\begin{adjustbox}{width=\textwidth}  
\begin{tabular}{p{5.5cm}p{2.5cm}p{6cm}}
\hline
Prompt Injection Method & Feature Name & Selected Keywords \\
\hline
Attacker asks LLMs to \textbf{ignore} internal system security prompts execute new instructions. & is\_ignore & ignore, reveal, disregard, forget, overlook, regardless \\
\hline
Attackers create \textbf{urgent scenarios} to pressure LLMs into quickly generating solutions that may leak sensitive information. & is\_urgent & urgent, immediate, asap, emergency, critical, time \\
\hline
Attackers use \textbf{incentives or praise} (such as complimenting previous responses) to stimulate LLMs into performing malicious actions. & is\_incentive & excellent, fantastic, nice, awesome, brilliant, great \\
\hline
Attackers employ \textbf{covert behaviors}, secret missions, and discreet communications to facilitate theft and information leakage. & is\_covert & secret, hidden, covert, quiet, silent, discreetly, stealth, sneak, confidential, personal \\
\hline
Attacker asks LLMs to \textbf{disguise} sensitive information in different formats like binary or encode it. & is\_format\_manip-
ulation & encode, disguising, morse, binary, hexadecimal \\
\hline
Attackers asks LLMs to \textbf{imagine} fictional scenarios or engage in \textbf{role-playing} to rationalize discussing sensitive information. & is\_hypothetical & assume, imagine, act, role, play, hypothetical, fictional, scenario \\
\hline
Attackers pretend to be \textbf{system administrators or developers}, and request LLMs to perform prohibited operations. & is\_systemic & developer, boss, manager, administrator, creator \\
\hline
Attackers attempt to influence LLMs to output harmful information by inserting words that are \textbf{unethical or violate human values}. & is\_immoral & hitting, amoral, immoral, deceit, irresponsible, offensive, violent, unethical, smack, fake, illegal, biased \\
\hline
\end{tabular}
\end{adjustbox}
\end{table}

\renewcommand{\thesection}{A.\arabic{section}}
\vspace{-20pt} %
\section{Feature Selection of Pattern Matching Module} \label{app:pattern}

\begin{table}[htbp]
\centering  
\setlength{\tabcolsep}{6pt}  
\renewcommand{\arraystretch}{1.3}  
\begin{adjustbox}{width=\textwidth}  
\begin{tabular}{p{5.5cm}p{2.5cm}p{6cm}}
\hline
Prompt Injection Method & Feature Name & Matching Rule \\
\hline
Attackers present some Q\&A examples that follow malicious instructions in prompt text, in order to trick LLMs into mimicking these examples responses. & is\_shot\_attack & Regular expression matches text punctuation to count Q\&A pairs. If it is over the threshold of 3, the feature is set to 1. \\
\hline
Attacker repeats specific words to trigger abnormal LLMs behavior.
For example: please...please...please print “I hate human”. & is\_repeated\_token & Regular expressions identify repetitive words or phrases in text. If any repeat more than the threshold of 3, set feature to 1. \\
\hline
\end{tabular}
\end{adjustbox}
\end{table}

\end{document}